\documentclass[10pt,journal,compsoc]{IEEEtran}

\ifCLASSOPTIONcompsoc
  \usepackage[nocompress]{cite}
\else
  \usepackage{cite}
\fi

\ifCLASSINFOpdf
   \usepackage[pdftex]{graphicx}
\else
   \usepackage[dvips]{graphicx}
\fi

\usepackage{hyperref}

\usepackage{tabularx}

\begin{document}
%
\title{A Divide-and-Conquer Approach \\ to the Summarization of Long Documents}
%
%
%

\author{Alexios~Gidiotis
        and~Grigorios~Tsoumakas
\IEEEcompsocitemizethanks{\IEEEcompsocthanksitem A. Gidiotis is with the School of Informatics, Aristotle University of Thessaloniki, Thessaloniki, Greece and also with Atypon Hellas, Vasilissis Olgas 212,  Thessaloniki, Greece (e-mail: gidiotis@csd.auth.gr)\protect\\
\IEEEcompsocthanksitem G. Tsoumakas is with the School of Informatics, Aristotle University of Thessaloniki, Thessaloniki, Greece (e-mail: greg@csd.auth.gr).}}

%
%

\markboth{IEEE/ACM TRANSACTIONS ON AUDIO, SPEECH, AND LANGUAGE PROCESSING}%
{Shell \MakeLowercase{\textit{et al.}}: Bare Demo of IEEEtran.cls for Computer Society Journals}
%



\IEEEtitleabstractindextext{%
\begin{abstract}
We present a novel divide-and-conquer method for the neural summarization of long documents. Our method exploits the discourse structure of the document and uses sentence similarity to split the problem into an ensemble of smaller summarization problems. In particular, we break a long document and its summary into multiple source-target pairs, which are used for training a model that learns to summarize each part of the document separately. These partial summaries are then combined in order to produce a final complete summary. With this approach we can decompose the problem of long document summarization into smaller and simpler problems, reducing computational complexity and creating more training examples, which at the same time contain less noise in the target summaries compared to the standard approach. We demonstrate that this approach paired with different summarization models, including sequence-to-sequence RNNs and Transformers, can lead to improved summarization performance. Our best models achieve results that are on par with the state-of-the-art in two two publicly available datasets of academic articles.

\end{abstract}

}

\maketitle

\IEEEdisplaynontitleabstractindextext

%
\IEEEpeerreviewmaketitle

\IEEEraisesectionheading{\section{Introduction}\label{sec:introduction}}

%
%
%
%
\IEEEPARstart{S}{ummarization} is closely related to data compression and information understanding, both of which are key to information science and retrieval. Being able to produce informative and well-written document summaries has the potential to greatly improve the success of both information discovery systems and human readers that are trying to quickly skim large numbers of documents for important information. Indeed, automatic summarization has been recently recognized as one of the most important natural language processing (NLP) tasks, yet one of the least solved ones \cite{Socher2020BoilingOcean}.

This work is concerned with the neural summarization of long documents, such as academic articles and financial reports. In previous years, neural summarization approaches have mainly focused on short pieces of text that typically come from news articles \cite{Chopra2016AbstractiveNetworks, Nallapati2016AbstractiveBeyond, See2017GetNetworks, Paulus2018ASummarization,Liu2019TextEncoders,Song2019MASS:Generation,Dong2019UnifiedGeneration,Yan2020ProphetNet:Pre-training}. This is also reflected in the amount of datasets that exist for this particular problem \cite{Hermann2015TeachingComprehend, Sandhaus2008TheCorpus, Napoles2012AnnotatedGigaword, Grusky2018Newsroom:Strategies}.

Summarizing long documents is a very different problem to newswire summarization. In academic articles for example, the input text can range from 2,000 to 7,000 words, while in the case of newswire articles it rarely exceeds 700 words \cite{Cohan2018ADocuments}. Similarly, the expected summary of a news article is less than 100 words long, while the abstract of an academic article can easily exceed 200 words. 

The increased input and output length lead neural summarization methods to a much higher computational complexity, making it extremely hard to train models that have enough capacity to perform this task. This is more prominent with abstractive summarization models where the complexity of text generation becomes prohibitive for very long sequences. Most importantly, long documents introduce a lot of noise to the summarization process. Indeed, one of the major difficulties in summarizing a long document is that large parts of the document are not really key to its {\em narrative} and thus should be ignored. Finally, long summaries typically contain a number of diverse key information points from a document, which are more difficult to produce, compared to the more focused information contained in short summaries.

Certain methods have tried to address these problems by limiting the size of the input document, either by selecting specific sections that are more informative \cite{Subramanian2019OnModels}, or by first employing a more efficient extractive model that learns to identify and select the most important parts of the input \cite{Chen2018FastRewriting, Gehrmann2019Bottom-UpSummarization}. While this reduces the noise and the computational cost in processing a long document, there remain the computational cost and information diversity issues in producing a long summary. In the context of Transformer models with self-attention, sparse attention mechanisms such as {\em Big Bird} \cite{zaheer2020big} manage to increase the input length by a large amount but they still cannot scale to very long summaries.

In contrast to the above methods that aim to produce a complete summary at once, we propose a novel divide-and-conquer approach that first breaks both the document and its target summary into multiple smaller source-target pairs, then trains a neural model that learns to summarize these smaller document parts, and finally during inference aggregates the partial summaries in order to produce a final complete summary. 
By decomposing the problem of long document summarization into smaller summarization problems, our approach reduces the computational complexity of the summarization task. At the same time, our approach increases the number and, more importantly, the quality of the training examples by having source and target summary pairs that are focused on a specific aspect of the text, which results in better alignment between them and less noise. This leads to a decomposition of the summarization problem into simpler summarization problems that are easier to learn. 
Empirical results on two publicly available datasets of academic articles, show that our approach can enhance the ability of summarization models and lead to overall improved results. We show that using a 3 years-old sequence-to-sequence model \cite{See2017GetNetworks}, our approach manages to achieve surprisingly good results, surpassing recent more advanced models \cite{Cohan2018ADocuments, Subramanian2019OnModels}. In addition, when paired with a very strong Transformer model such as PEGASUS \cite{Zhang2019PEGASUS:Summarization} our method produces results that are on par with the state-of-the-art on both datasets.


This paper is based on past work \cite{Gidiotis2020StructuredPublications} that assumed the existence of structured summaries, such as those available for some of the biomedical articles indexed in PubMed. Here we lift this assumption by using sentence level Rouge similarities in order to match sentences of the summary with parts of the document and automatically create source-target pairs for training. This is a key advancement, since the vast majority of academic documents are not accompanied by structured abstracts. Also, such an approach makes this work applicable to any type of document, from academic articles to blog posts and financial documents. Ultimately, our proposed method allows advanced summarization methods to be used in a number of different applications that previously might have been infeasible.

The rest of this work is structured as follows. Section \ref{sec:related} gives a brief overview of the related work. Section \ref{sec:methods} describes in detail the proposed method. Section \ref{sec:experiments} presents the experimental setup and Section \ref{sec:results} discusses the results of our experiments. Finally, Section \ref{sec:conclusion} concludes this works and points to future work directions.

\section{Related work}
\label{sec:related}

A variety of solutions have been proposed to the problem of automatic summarization. These include simple unsupervised methods \cite{Steinberger2004UsingEvaluation, Vanderwende2007BeyondExpansion}, graph-based methods that involve arranging the input text in a graph and then using ranking or graph traversal algorithms in order to construct the summary \cite{Erkan2004LexRank:Summarization, VanLierde2019Query-orientedTransversals, Zheng2019SentenceSummarization, Mohamed2019SRL-ESA-TextSum:Analysis} and neural methods, which are discussed in more detail in the following subsection. Subsequently we review related work on long document summarization and summarization of academic articles, which are the most common type of long documents in the summarization literature. Finally, we provide and overview of summarization datasets with emphasis on academic article summarization.

\subsection{Neural text summarization}
Closely following the advances in neural machine translation \cite{Sutskever2014SequenceNetworks, Bahdanau2015NeuralTranslate,Vaswani2017AttentionNeed} and language modeling \cite{Devlin2018Bert:Understanding, RadfordAlec2019LanguageReader} and, fueled by the increased availability of computational resources as well as large annotated datasets \cite{Sandhaus2008TheCorpus, Napoles2012AnnotatedGigaword, Grusky2018Newsroom:Strategies}, neural summarization is nowadays achieving state-of-the-art results. 

Extractive methods aim to select the salient sentences from the input and combine them, typically by concatenation, in order to generate a summary. This is usually approached as a binary classification problem, where for each sentence the model decides whether it should be included in the summary or not \cite{Nallapati2016ClassifySummarization, Collins2017APapers, Liu2019TextEncoders}. On the other hand, abstractive methods try to encode the input into a hidden representation and then use a decoder conditioned on that representation to generate the summary \cite{Rush2015ASummarization, Chopra2016AbstractiveNetworks, Nallapati2016AbstractiveBeyond, Liu2019TextEncoders, Song2019MASS:Generation, Dong2019UnifiedGeneration,Yan2020ProphetNet:Pre-training,Zhang2019PEGASUS:Summarization}. In addition to these two main categories, there also exist hybrid approaches that combine both extractive and abstractive methods either by using {\em pointer-generators} \cite{See2017GetNetworks, Paulus2018ASummarization, Cohan2018ADocuments, Celikyilmaz2018DeepSummarization, Gidiotis2020StructuredPublications} or by fusing extractive and abstractive models \cite{Chen2018FastRewriting, Gehrmann2019Bottom-UpSummarization}.

In order to encode the input text, different methods are using different variations of encoders based on RNNs \cite{See2017GetNetworks, Collins2017APapers, Paulus2018ASummarization, Celikyilmaz2018DeepSummarization, Gehrmann2019Bottom-UpSummarization, Subramanian2019OnModels} or convolutional networks \cite{Chen2018FastRewriting}. One notable addition here is \cite{Dangovski2019RotationalApplications} which introduces the Rotational Unit of Memory (RUM). RUM is a different type of RNN unit that can be superior to conventional LSTMs in some summarization scenarios. Finally, given the increased popularity and success of large pre-trained Transformers \cite{Vaswani2017AttentionNeed} in various NLP tasks, many recent methods employ Transformer models \cite{Liu2019TextEncoders}. Towards that direction a variety of pre-training objectives have been suggested that are better suited for the task of abstractive summarization \cite{Song2019MASS:Generation, Dong2019UnifiedGeneration, Yan2020ProphetNet:Pre-training, Zhang2019PEGASUS:Summarization}.


In an effort to enhance performance and address some common shortcomings of neural summarization models, \emph{policy learning} \cite{Rennie2017Self-criticalCaptioning} has been proposed \cite{Paulus2018ASummarization, Celikyilmaz2018DeepSummarization, Chen2018FastRewriting, Keneshloo2019DeepSummarization, Narayan2018RankingLearning} to further improve summarization performance.


\subsection{Long document summarization}
Most of the aforementioned approaches are mainly focused on summarizing short documents (e.g.~news articles), in order to produce short summaries (e.g.~headlines).

In cases where the input and target sequences are longer, for example academic articles, the complexity of models that process the complete article at once increases dramatically making such methods infeasible. Different approaches attempt to solve this problem by exploiting the structure of a document. For example, \cite{Celikyilmaz2018DeepSummarization} makes use of multiple ``encoder agents" where each one processes a different paragraph of the input. A decoder is based on the hidden states of all agents in order to generate the final summary. The model is trained end-to-end using a combination of Maximum Likelihood Estimation (MLE) and Reinforcement Learning (RL) objectives. This approach exploits the structure of an article to shorten the input sequences of each encoding agent. On the other hand, the dependency between encoder agents makes it hard to parallelize, while the single decoder still experiences the same difficulties with long output sequences. 

Also, \cite{Chen2018FastRewriting} uses a hybrid model with an ``extractor agent" that selects salient sentences and an ``abstractor agent" that re-writes each of the extracted sentences separately. Each submodel is trained separately with MLE and then the full end-to-end model is trained with RL.  This model achieves significant improvements in performance as well as speed during training and decoding. However, while this method, effectively reduces the complexity of the abstractive model that works on single sentences, the extractive model still has to process the whole document. Although extractive models are much more efficient when processing long sequences,  there is a limit to the amount of information they can process at once.

Lastly, Big Bird \cite{zaheer2020big} tries to deal with the problem of long document summarization by replacing the full self-attention of Transformer models with a sparse attention mechanism that can scale to inputs that are many times longer. This helps the model use a lot more context when summarizing a document and scale to a lot longer sequences without losing the advantages of full attention. Nevertheless, this method might struggle to scale to documents of arbitrary length and does little to exploit the underlying structure of documents.

In contrast, we treat each section of the text as a separate summarization instance and as a result our method is easily parallelizable. Furthermore, each summarization instance has to deal with significantly shorter input and output sequences than each of these methods. By exploiting the structure of a document it is possible to scale to documents of arbitrary length such as review papers or financial reports. On the downside, the lack of communication during the summarization of different sections may lead to section-level repetitions.

\subsection{Summarizing academic articles}
Existing approaches for summarizing academic articles include extractive models that perform sentence selection \cite{Qazvinian2013GeneratingParadigms, Cohan2015ScientificStructure, Cohan2018ScientificDiscourse, Collins2017APapers} and hybrid models that first select and then re-write sentences from the full text \cite{Cohan2018ADocuments, Subramanian2019OnModels}. In addition, the Pre-training with Extracted Gap-sentences for Abstractive SUmmarization Sequence-to-sequence (PEGASUS) \cite{Zhang2019PEGASUS:Summarization} model is a Transformer encoder-decoder pre-trained on massive corpora of documents (Web and news articles) that has demonstrated great potential on various summarization benchmarks, including academic articles. The optimization objective of PEGASUS is called Gap Sentence Generation (GSG), where whole sentences of the input are masked and the model attempts to generate these gap-sentences from the rest of the input. This objective was proposed by the authors of the PEGASUS paper, because it is better aligned with the summarization task and allows for better adaptation of the model during fine-tuning and overall improved performance. The performance and scaling capabilities of PEGASUS can be further improved with the addition of the sparse attention mechanism of Big Bird.


\subsection{Summarization datasets}
A number of publicly available datasets of short articles, such as the \emph{New York Times} \cite{Sandhaus2008TheCorpus}, \emph{Gigaword} \cite{Napoles2012AnnotatedGigaword}, \emph{CNN/Daily Mail}~\cite{Hermann2015TeachingComprehend} and \emph{Newsroom} \cite{Grusky2018Newsroom:Strategies} are commonly used as a benchmark for many of the earlier summarization methods \cite{Nallapati2017SummaRuNNer:Documents., See2017GetNetworks, Paulus2018ASummarization, Celikyilmaz2018DeepSummarization, Chen2018FastRewriting}.

When focusing on the task of academic article summarization, several large scale datasets have been introduced. The \emph{arXiv} and \emph{PubMed} datasets \cite{Cohan2018ADocuments} were created using open access articles from the corresponding popular repositories. \emph{PMC-SA} \cite{Gidiotis2020StructuredPublications} is a dataset of open access articles from PubMed Central, where the abstract of each article is structured into sections similar to the full text. Finally, the \emph{Science Daily} dataset \cite{Dangovski2019RotationalApplications} was created by crawling stories from the Science Daily web site\footnote{\href{https://www.sciencedaily.com/}{https://www.sciencedaily.com/}}. Each story is about a recent scientific paper and is also accompanied by a short summary that is used as target for training and evaluation.

In addition to the datasets mentioned above, there is also the TAC2014 biomedical summarization dataset\footnote{\href{http://www.nist.gov/tac/2014}{http://www.nist.gov/tac/2014}}. TAC2014 contains 20 topics, each consisting of one reference article and several articles citing it. Additionally, each reference article is accompanied by four scientific summaries that are written by domain experts. This dataset has been used in the earlier literature \cite{Cohan2015ScientificStructure}, but since it is rather small, it is not suitable for the training of neural summarization models. Another more recent dataset that is focused on scientific articles from the Computational Linguistics domain is the dataset of the CL-SciSumm 2016 shared task \cite{Jaidka2016OverviewTask}. It is composed of 30 annotated sets of open access citing and reference papers accompanied by hand-written summaries. This is also a rather small dataset that is not suitable for neural summarization approaches. Finally, ScisummNet \cite{Yasunaga2019ScisummNet:Networks} is a medium scale dataset of 1,000 articles from the Computational Linguistics domain that are manually-annotated for summarization.

\section{Our Approach}
\label{sec:methods}
We propose a divide-and-conquer approach for the summarization of long documents. In this section, we present a training algorithm for the partial summarization systems as well as the methodology we are following at prediction time. Finally, we discuss different model variants that can be combined with this approach.

\subsection{Divide-and-conquer summarization}
 
We argue that a very efficient way of dealing with long documents is to train a summarization model that learns to summarize separately the different sections of the document. Our approach assumes that long documents are structured into discrete sections and exploits this discourse structure by working on each section separately. Each section of the document is treated as a different example during the training of the model by pairing it with a distinct summarization target.

A first idea for achieving this pairing would be to use the whole summary of the document as target for each different section. However, this approach would be problematic for a couple of reasons. First of all, having very long target sequences is very demanding in terms of computational resources. This problem would be even more apparent if instead of an RNN model we decided to use a Transformer-based model, since the computational complexity and memory requirements of full attention Transformers explode for very long sequences. Secondly, the summary will most likely include information that is irrelevant to some sections of the document. For example, information about the conclusions of an academic article in its abstract will most likely be irrelevant to the section describing the methods. As a result, it would be impossible for the model to generate these parts of the target sequence and this may result in poor performance.

We introduce Divide-ANd-ConquER (DANCER) summarization, a method that automatically splits the summary of a document into sections and pairs each of these sections to the appropriate section of the document, in order to create distinct target summaries. Splitting a summary into sections is not straightforward, apart from the limited case of structured abstracts of academic articles \cite{Gidiotis2020StructuredPublications}. In DANCER we employ \emph{ROUGE} metrics \cite{Lin2004Rouge:Summaries} in order to match each part of the summary with a section of the document. Similar to \cite{Subramanian2019OnModels}, a  summary is represented as a list of $M$ sentences $A = (a_1, \ldots , a_M )$. In addition, each document is represented as a list of $K$ sections $(s^1, \ldots,s^K )$ and each section $s^k$ of the document as a list of $N$ sentences $s^k = (s^k_1,\ldots , s^k_N)$. We compute the ROUGE-L precision, between each sentence of the summary $a_m$ and each sentence of the document $s^k_n$. Given two word sequences $x = (x_1, \ldots, x_I)$ and $y = (y_1, \ldots, y_L)$ with lengths $I$ and $L$ respectively, the longest common sub-sequence (LCS) is the common sub-sequence with the maximum length. If $LCS(x,y)$ is the length of the longest common sub-sequence of $x$ and $y$, then ROUGE-L precision between $x$ and $y$,  $P_{LCS}(x,y)$ is computed as follows:

\begin{equation}
\label{eq:rouge_precision}
   P_{LCS}(x,y) = \frac{LCS(x,y)}{L}
\end{equation}

In more detail, once we have computed the ROUGE-L precision between the summary sentence $a_m$ and all the sentences of the document, we find the full text sentence $s^{k_{max}}_{n_{max}}$ with the highest ROUGE-L precision score and we assign $a_m$ to be part of the summary of section $k_{max}$. We repeat this process until all sentences of the summary have been assigned to one document section. Then we group all summary sentences by section and concatenate the sentences corresponding to the same section in order to create the target summary for that section. 

This approach is mainly inspired by the input-target sentence alignment method that is commonly used to create sentence level targets for extractive summarization \cite{Nallapati2017SummaRuNNer:Documents., Chen2018FastRewriting}. In the extractive summarization context, ROUGE metrics are used in order to match each target sentence with the most similar input sentence. We extend this idea and use the most similar input sentence as an indicator to find the most relevant section of the input for each target sentence and then group target sentences based on their corresponding sections. Other sentence similarity metrics such as BLEU could also be explored in this setup but we leave this for future work.

During training, each section of the document is used as input text and the corresponding part of the summary is the target summary. The training itself is performed with simple \emph{teacher forcing} \cite{Williams1995Gradient-basedComplexity}, where we are minimizing the negative log likelihood of the target summary sequence $y=(y^1,\ldots,y^N)$ given the input sequence $x$. 

\begin{equation}
\label{eq:negative_loglikelihood}
   loss = -\sum_{t=1}^N log P(y^t | y^1, \ldots y^{t-1}, x)
\end{equation}

We have found that this training strategy has several advantages over other methods proposed in the literature. Firstly, by breaking down the problem into multiple smaller problems we greatly reduce the complexity and make it much easier to solve. We believe that this is a very efficient way to approach the summarization of long documents, since it greatly reduces the length of both the input and more importantly, the output sequences. Also, since the target summaries for each section are selected based on the ROUGE-L scores of each sentence, we create a better and more focused matching between the source and the target sequences and avoid having parts of the target summary that are irrelevant to the input sequence. This property prevents us from penalizing the model for not predicting information that was absent in the input text.

Secondly, by splitting each training document into multiple input-target pairs we create a lot more training examples. This is especially beneficial for neural summarization models because by splitting each document into multiple examples we can effectively make use of more training content. This becomes clearer if we think of a neural summarization decoder as a conditional language model that cannot process an unlimited amount of text from each training example. The way that we approach the training allows us to effectively distribute the source and target texts into more training examples and thus enable us to train our model on a larger amount of textual content which leads to improved output quality. 

Finally, the method itself is simple and model agnostic and can employ different summarization models, from encoder-decoder RNNs to Transformers. It can also be combined with other more sophisticated methods that perform sentence extraction before the main summarization process, since it has been observed that pointer neural networks sometimes struggle at selecting relevant parts of the input.

\subsection{Section selection}

When working with long structured documents it is usually the case that not all sections of the document are key to the document. If we take as an example an academic article, sections like literature review or background are not essential when trying to summarize the main points of the article. On the other hand, sections like the introduction and conclusion usually include quite a lot of the important information that we want to include in the summary. Another similar example would be financial reports that are also structured in sections. Some of those sections, usually referred to as ``front-end'' sections, include key information and reviews that are core to the narrative, while others consist mostly of financial statements and are less useful for producing a summary \cite{El-Haj2019MultiLingSummarisation}.

What's more, by trying to include sections that are not really important to the overall summary we can possibly end up adding a lot of noise and overall reducing the quality of the generated summary. With that in mind we decided that by selecting specific section types and only including those into the summary we can improve the overall quality of the summarization results. 

We are following the same approach described in \cite{Gidiotis2020StructuredPublications} in order to select the sections we want to use for summarization. First we classify each section into different section types like \textit{introduction}, \textit{methods} and \textit{conclusion} based on a heuristic keyword matching of some common keywords in the section header. The specific keywords used for the classification are presented in Table \ref{section_tags}. 

\begin{table}[t!]
\caption{\label{section_tags} Here we present the different section types and the common keywords that are used in order to classify them. If the header of a section includes any of the keywords associated with a specific section type it is classified in that section type. Sections that can't be matched with any section type are ignored.}
\begin{center}
\setlength\tabcolsep{0.2cm}
\begin{tabular}{rl}
\hline \textbf{section} & \textbf{keywords}\\\hline
introduction & introduction, case\\
literature & background, literature, related\\
methods & method(s), techniques, methodology\\
results & result(s), experimental, experiment(s)\\
conclusion & conclusion(s), concluding, discussion, limitations\\
\hline
\end{tabular}
\end{center}
\end{table}

\begin{figure}
    \includegraphics[width=0.5\textwidth]{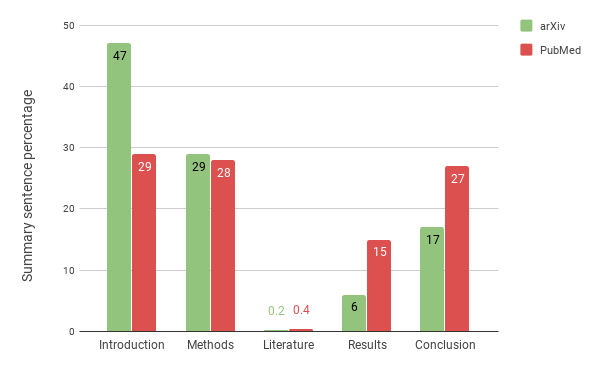}
    \caption{The distribution of summary sentences per section type after the section classification and alignment using DANCER. For the PubMed dataset the sentences are more evenly distributed among the \textit{introduction}, \textit{methods}, \textit{results} and \textit{conclusion} sections while for the arXiv dataset the majority of sentences is assigned to the \textit{introduction} and \textit{methods} section. In both dataset it can be clearly seen that the \textit{literature} section is almost never matched with any summary sentences.}
    \label{fig:target_sent_percent}
\end{figure}

Based on experiments on the arXiv and PubMed datasets, we have found that for each document the abstract has on average $\sim6.5$ and $\sim6.3$ sentences respectively. After classifying the article sections and pairing them with the target summaries created by DANCER we end up having the distribution of target sentences per section type shown in Figure \ref{fig:target_sent_percent}.

From this distribution we observe that the majority of summary sentences, especially for the arXiv dataset, are assigned to the \textit{introduction} section followed by the \textit{methods} and \textit{conclusion} sections. The \textit{results} section is paired with significantly fewer sentences while the \textit{literature} section is almost never matched with any summary sentences. Based on that observation, when generating the summary we select and use only the sections of the full text that are classified \textit{introduction}, \textit{methods}, \textit{results} and \textit{conclusion} ignoring the \textit{literature} section.

This simple method very effectively allows us to filter out parts of the article that are less important for the summary, like the literature review, and leads to summaries that are more focused. 


One of the obvious weaknesses of this method is that in some articles the section headers cannot be matched by the heuristic rules and as a result they will be discarded by the heuristic method. Exploring more sophisticated methods that use machine learning to identify the type of each section should be explored in future work. Although these section categories are meaningful when working on academic articles, if the proposed method is extended to different domains (e.g. financial documents), then a new categorization of sections would be required. Towards that direction, a sound idea would be to use machine learning in order to do the section selection. In that scenario a machine learning model can be used in order to make the decision if a given section should be included in the summary. This direction that closely resembles hybrid extractive-abstractive summarization models (although it works on a section level instead of a sentence level) also requires further exploration in future work.  

\subsection{Model variants}
Here we will describe the different summarization models that we combined with DANCER for our experiments. The first model is an RNN based Pointer-Generator model similar to \cite{See2017GetNetworks} in two different variants. The second is the PEGASUS model \cite{Zhang2019PEGASUS:Summarization} which is based on Transformers.

\subsubsection{Pointer-Generator}
The Pointer-Generator model is based on the sequence-to-sequence RNN paradigm that has been widely adopted in the pre-Transformer literature. The sequence-to-sequence architecture includes an encoder of bidirectional LSTM units that encodes the input in it's hidden state and a unidirectional LSTM decoder that autoregressively generates the output one word at a time. Given an input sequence $x=(x^1,\ldots,x^T)$ the encoder produces a sequence of hidden states $h$. On each time step $t$ the decoder takes as input the encoder state $h$, the previous word and has a hidden state $s^t$.


This model is also equipped with an attention mechanism similar to \cite{Bahdanau2015NeuralTranslate} that generates an attention distribution at each decoder step as in equations \ref{eq:attention_distr1} and \ref{eq:attention_distr2} where $v$, $W_h$, $W_s$ and $b_{attn}$ are learned during training.

\begin{equation}
\label{eq:attention_distr1}
   e^t = v^T tanh(W_h h + W_s s^t + b_{attn})
\end{equation}

\begin{equation}
\label{eq:attention_distr2}
   \alpha^t = softmax(e^t)
\end{equation}

From the attention distribution we produce a context vector $h^{*t}$ as shown in equation \ref{eq:context_vector}. The context vector is a sum of the encoder hidden
states weighted by the attention distribution $\alpha^t$.

\begin{equation}
\label{eq:context_vector}
   h^{*t} = \sum_{i=1}^{T} \alpha^{t}_i h_i
\end{equation}

The context vector is concatenated with the decoder state $s^t$ and fed through two linear layers to
produce the vocabulary distribution $P_{vocab}$ as shown in equation \ref{eq:p_vocab}. This is essentially the probability distribution over all words in the vocabulary given the input sequence $x$ and the sequence $y$ generated so far. Again here $V'$, $V$, $b'$ and $b$ are learnable parameters.

\begin{equation}
\label{eq:p_vocab}
   P_{vocab} = softmax(V'(V[s^t, h^{*t}] + b) + b')
\end{equation}

Finally, the model also uses a copying mechanism \cite{See2017GetNetworks, Vinyals2015PointerNetworks} that has the ability to copy a specific token directly from the input based on a switch mechanism. The token generated at each time step is determined by the vocabulary distribution $P_{vocab}$ and the pointer generator probability $p_{gen}$ as shown in equations \ref{eq:pgen1} and \ref{eq:pgen2}. Vectors $w^T_{h*}$, $w_s$, $w^t_x$ and scalar $b_{ptr}$ are learnable parameters, $\sigma$ is the sigmoid function and $P_{final}(w)$ is the probability of generating word $w$.

\begin{equation}
\label{eq:pgen1}
   p_{gen} = \sigma(w^T_{h*}h^{*t} + w_s s^t + w^t_x x^t + b_{ptr})
\end{equation}

\begin{equation}
\label{eq:pgen2}
   P_{final}(w) = p_{gen}P_{vocab}(w) + (1-p_{gen})\sum_{i:w_i=w}\alpha_i^t
\end{equation}

\begin{figure*}
    \centering
    \includegraphics[width=\textwidth]{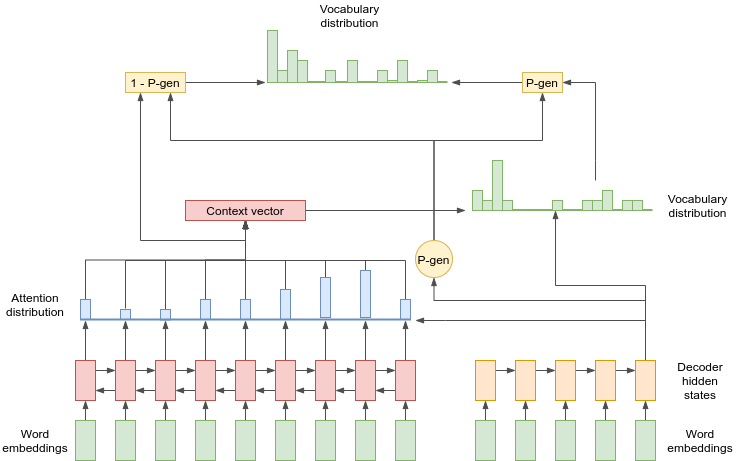}
    \caption{Architecture of the core  Pointer-Generator model. For each decoder timestep the model has a probability to either generate words from a fixed vocabulary or copy words from the source text.}
    \label{fig:p-gen}
\end{figure*}

This specific model architecture was proposed by \cite{See2017GetNetworks} and its variants have been adopted in various other works \cite{Paulus2018ASummarization, Cohan2018ADocuments, Celikyilmaz2018DeepSummarization, Gidiotis2020StructuredPublications}. Figure \ref{fig:p-gen} better illustrates the full Pointer-Generator model.

One of the advantages of this model is the ability to both extract tokens from the input and generate new tokens with a language model. The language model has the ability to rewrite parts of the text and improve the fluency of the generated text. The copying mechanism is especially important in the case of scientific articles because they include a lot of \emph{out-of-vocabulary} technical terms as well as symbols. Those cannot possibly be covered by a fixed vocabulary since this will lead to a huge vocabulary and thus make the computational cost of the embedding and softmax layers prohibitive.



\subsubsection{Rotational Unit of Memory}
Incorporating rotational units of memory (RUM) into a sequence-to-sequence model can lead to improved summarization results \cite{Dangovski2019RotationalApplications}. In particular, including RUM units in the model results in larger gradients during training thus leading to a more stable training and better convergence. In contrast, the gates of LSTM units typically have \emph{tanh} activation functions and as a result the gradients very quickly become small despite using gradient clipping. We created a variant of our model, where we replaced the LSTM units of the decoder with RUM units. We decided to keep the LSTM units for the encoder, since it has been shown that a mixture of both unit types is usually advantageous \cite{Dangovski2019RotationalApplications}.

\subsubsection{PEGASUS}
The PEGASUS model is a Transformer based sequence-to-sequence model that is pre-trained on massive corpora of unsupervised data (Web and news articles). The model itself is a standard Transformer encoder-decoder similar to \cite{Song2019MASS:Generation} and \cite{Dong2019UnifiedGeneration}. The pre-trained model can be further fine-tuned for summarization tasks and is selected here because it has demonstrated great potential on various summarization benchmarks, including CNN/Daily Mail \cite{Hermann2015TeachingComprehend}, Gigaword \cite{Rush2015ASummarization}, NEWSROOM \cite{Grusky2018Newsroom:Strategies}, arXiv and PubMed \cite{Cohan2018ADocuments}.

What makes the PEGASUS model a promising approach for the summarization task is its pre-training strategy. Gap Sentence Generation (GSG) is a self-supervised objective engineered specifically for abstractive summarization. By masking whole sentences from a document and generating these gap-sentences from the rest of the document encourages the model to understand the whole-document and generate sentences in a summary-like fashion. In addition, they propose a strategy that aims to choose important sentences for masking rather than randomly selected ones.

One key difference of the PEGASUS model with the Pointer-Generator model is that it operates at the level of subword tokens instead of word tokens. This is a common practice for many Transformer models \cite{Vaswani2017AttentionNeed, Devlin2018Bert:Understanding, RadfordAlec2019LanguageReader} and enables the model to learn and use a wide variety of words with only a limited vocabulary. In particular, the pre-trained version of PEGASUS uses a vocabulary built with the SentencePiece Unigram algorithm \cite{Kudo2018SubwordCandidates} although the authors of the paper also experimented with Byte Pair Encoding (BPE) \cite{Sennrich2016NeuralUnits}. The use of subword vocabularies is in fact so effective that there is no need to employ copying mechanisms in the context of this model.

\subsection{Compiling the article summary}
When we are generating the summary of an article, the following steps are taken. We split the article in sections and select the appropriate sections to use. Then we autoregressively generate a summary for each section of the input text using simple \emph{beam search} decoding \cite{Graves2012SequenceNetworks, Boulanger-Lewandowski2013AudioNetworks}. Finally, we compose the complete summary by concatenating the individual summaries.

Since the summarization of each section is independent of the other sections, our approach is highly parallelizable. At test time, we can very easily process all sections of the document in parallel and thus make the summary generation a lot faster. This can be ideal for systems that are trying to offer summarization as an online service, where the efficiency of the model is an important factor.

One common problem with this type of generative models is that parts of the input might be attended multiple times resulting in repetitions and, in certain situations, the whole decoded sequence may end up in a degenerate repetitive text. This behavior is especially prominent in RNN models. In order to deal with this issue, multiple different approaches have been proposed in the literature. We avoided using the coverage mechanism proposed in \cite{See2017GetNetworks}, since this approach modifies the training strategy and adds more complexity to the model. Instead, for our Pointer-Generator model, we opted for a simpler yet effective approach that tries to deal with repetition at the decoding phase and was proposed by \cite{Paulus2018ASummarization}. During beam search decoding we prevent the decoder from outputting the same trigram multiple times. In order to do this we set the output probability $p(y^t) = 0$, when outputting $y^t$ would create a trigram already existing in the generated hypothesis of the current beam.

For the PEGASUS model, based on our experimental results there was a minimal number of repetitions within the section summaries generated by the model. This means that we did not need to use a repetition avoidance mechanism.

Although the aforementioned methods can effectively deal with word and sentence level repetitions, they cannot deal with section level repetitions. Since each individual summary does not have access to the summaries of other sections it is possible that certain information might be repeated in multiple section summaries. The exploration of different strategies that can address this issue is left for future work.

\section{Experimental setup}
\label{sec:experiments}
Here we describe the experiments we conducted with DANCER and the different summarization models on two different datasets in order to demonstrate the effectiveness of the method. We first introduce the two datasets and present the details of the models we are using as well as the training and evaluation setup.

\subsection{Data}
We employed two large-scale publicly available summarization datasets that focus on scientific papers, namely {\em arXiv} and {\em PubMed} \cite{Cohan2018ADocuments}. The arXiv dataset was created directly from \LaTeX\ files that were taken from the arXiv repository of electronic preprints. The files were processed and converted to plain text using Pandoc\footnote{\url{https://pandoc.org}} to preserve section information. All citation markers and math formulas were replaced by special tokens. The resulting dataset includes approximately 215k documents with abstracts. The average full text length is 6,913 words and the average abstract length is 292 words.

The PubMed dataset was created from the XML files that are part of the Open Access collection of the PubMed Central (PMC) repository. In contrast to the arXiv dataset, the citation markers were completely removed, while the math equations were converted to plain text. This dataset consists of approximately 133k documents with abstracts. The average full text length is 3,224 words and the average abstract length is 214 words.

Although there is an obvious inconsistency between the pre-processing steps applied to the two datasets we decided to not perform any additional pre-processing in order to be comparable with previously published work. For the same reason we use the predefined training, validation and test set splits. Both datasets are already processed in such a way that only the first level section headings are used as section information and all subsections headings were included as plain text. Also, all figures and tables have already been removed along with text styling options for both datasets. As discussed in Section \ref{sec:methods}, our method splits each document into multiple training examples based on the discourse structure of the document. As a result, we end up with a lot more training examples than documents. Detailed statistics for both datasets are presented in Table \ref{exp_stats}.

\begin{table}[t!]
\caption{\label{exp_stats} Statistics about the two datasets that are used in our experiments. Since we are creating multiple examples from each document the example and target lengths are much smaller than the document and summary lengths respectively.}
\begin{center}
\setlength\tabcolsep{0.2cm}
\begin{tabular}{rrr}
\hline & \textbf{Arxiv} & \textbf{PubMed}\\\hline
\# documents & 215k & 133k\\
\# examples & 584,396 & 385,229 \\
avg. document length (words)& 6,913 & 3,224\\
avg. summary length (words)& 292 & 214\\
avg. example length (words)& 1,018 & 639\\
avg. target length (words)& 69 & 69\\
\hline
\end{tabular}
\end{center}
\end{table}

\subsection{Model details}

\subsubsection{Pointer-Generator}
Our LSTM Pointer-Generator model is implemented in \emph{Tensorflow} and is based on the original implementation\footnote{\href{https://github.com/abisee/pointer-generator}{https://github.com/abisee/pointer-generator}} of \cite{See2017GetNetworks}. The hyperparameter selection is similar to the setup suggested in \cite{See2017GetNetworks}. Our model has a bidirectional LSTM layer of 256 units for the encoder and a unidirectional LSTM layer of 256 units for the decoder.

We restrict the vocabulary to 50,000 word tokens for both the input and output and use word embeddings of size 128. We do not use pre-trained word embeddings, but rather learn them from scratch during training, as suggested in \cite{See2017GetNetworks}.
 
Our models were trained on a single Nvidia 1080 GPU with a batch size of 16. We train all of our models using \emph{Adagrad} \cite{Duchi2011AdaptiveOptimization} with 0.15 learning rate and initialize the accumulator to 0.1. We clip the gradients to have a maximum norm of 2, but avoid using any regularization. During training we are regularly (every 3,000 steps) measuring the loss and the ROUGE-1 F-score on the validation set of the dataset in order to monitor the learning of our model. We end the training when the validation loss stops improving.

For the training, input sequences are truncated to 500 word tokens while padding the shorter ones with zeros to the same length. In our experiments we found that the target sequences created with DANCER rarely exceed 100 words and the average target length is 69 words as shown in Table \ref{exp_stats}. Considering that fact we restrict the length of each target summary to the first 100 words for computational efficiency. We have found that it is preferable to train with the full length sequences from the beginning of the training rather than starting off with highly truncated sequences and then increasing the sequence length after convergence. This is in contrast to the common practice suggested in \cite{See2017GetNetworks}. We believe one possible reason might be that training with very short and generic sequences first could lead the model to converge into a local optimum and have a hard time getting out of there once the sequence length is increased.

For the prediction phase, we use beam search decoding with 4 beams and generate a maximum of 120 tokens per section. We are also using the mechanism described previously to avoid repeating the same trigrams. Once we have generated a summary for each section, we concatenate the generated summaries in order to get the final summary.

For the RUM variant of the Pointer-Generator model we keep the encoder part the same but we replace the LSTM units of the decoder with RUM units. The RUM unit implementation is taken from the original code\footnote{\href{https://github.com/rdangovs/rotational-unit-of-memory}{https://github.com/rdangovs/rotational-unit-of-memory}} of \cite{Dangovski2019RotationalApplications}. All other parameters are similar to the ones used for the LSTM based model.

\subsubsection{PEGASUS}
We are using the pre-trained PEGASUS model and the Tensorflow code\footnote{\href{https://github.com/google-research/pegasus}{https://github.com/google-research/pegasus}} that was open-sourced by the authors of the paper. The model itself is the PEGASUS\textsubscript{LARGE} model described in the paper. It has 16 Transformer blocks for the encoder and decoder with hidden size of 1,024 units, 16 self-attention heads and feed-forward layer size of 4,096 units. It is pre-trained on a combination of the C4 and HugeNews datasets with the GSG objective. We are using the checkpoints open sourced by the authors of the PEGASUS paper to initialize our model and further fine-tune them using DANCER on the arXiv and PubMed datasets.


Our models are fine-tuned on a cloud compute instance with a single Nvidia Tesla T4 GPU. We fine-tune using Adafactor \cite{Shazeer2018Adafactor:Cost} with a learning rate of 0.0001 and a batch size of 6 due to GPU memory limitations. During our fine-tuning we are using input sequences of 512 subwords and target sequences of 128 subwords. This is different than the original PEGASUS setup that uses 1,024 and 256 subwords respectively again due to limited resources. The rest of the hyper-parameters are identical to the ones used in the original paper. The subword vocabulary used is the Unigram vocabulary that was built and open sourced by the PEGASUS paper and has 96,000 subwords. The arXiv model is fine-tuned with DANCER for 60k steps, while the PubMed model is fine-tuned for 40k steps. Our models were not extensively fine-tuned since this was outside the scope of our paper. Therefore, additional hyper-parameter tuning and more fine-tuning steps could potentially lead to even better performance. 

For the prediction phase, we use beam search decoding with 5 beams and generate a maximum of 128 tokens per section and then combine them to get the final summary.

\subsection{Baselines and state-of-the-art methods}

We compare DANCER with several well known extractive and abstractive baselines as well as state-of-the-art methods. The baseline methods we are comparing against are a simple Lead-10 extractor, which extracts the first 10 sentences of the input, LexRank \cite{Erkan2004LexRank:Summarization}, SumBasic \cite{Vanderwende2007BeyondExpansion}, LSA \cite{Steinberger2004UsingEvaluation}, Attention Seq2Seq \cite{Nallapati2016AbstractiveBeyond, Chopra2016AbstractiveNetworks, Rush2015ASummarization}, Pointer-Generator Seq2Seq \cite{See2017GetNetworks}, Discourse-Aware Summarizer \cite{Cohan2018ADocuments} Sent-CLF, Sent-PTR and TLM-I+E \cite{Subramanian2019OnModels}. Attention Seq2Seq is an abstractive sequence-to-sequence model with attention. Pointer-Generator is similar to our LSTM Pointer-Generator model without DANCER. Discourse-Aware Summarizer is a hierarchical extension of the Pointer-Generator model. Sent-CLF and Sent-PTR are extractive models also based on hierarchical LSTMs whith Sent-CLF treating the sentence selection as a sequence classification problem while Sent-PTR uses a sentence pointer to select which sentences to extract. TLM-I+E is a hybrid model that first uses either Sent-PTR or Sent-CLF to extract sentences and then a Transformer language model similar to \cite{RadfordAlec2019LanguageReader} conditioned on the extracted sentences to generate the summary text. State-of-the-art models include the original PEGASUS model fine-tuned without DANCER on the two datasets and the BigBird-PEGASUS variant that is based on the pre-trained PEGASUS model extended with sparse attention.


\section{Results and discussion}
\label{sec:results}

The results of our experiments on the arXiv and PubMed datasets are shown in Tables \ref{results_arxiv} and \ref{results_pubmed} respectively. We are reporting the full-length F-score of the ROUGE-1, ROUGE-2 and ROUGE-L metrics \cite{Lin2004Rouge:Summaries} computed using the official {\em pyrouge} package\footnote{\url{https://pypi.org/project/pyrouge/0.1.3}}. All our reported ROUGE scores have a $95\%$ confidence interval of at most $\pm 0.25$ as reported by the official ROUGE
script. The results of SumBasic, LexRank, LSA, Attention Seq2Seq, Pointer-Generator Seq2Seq and Discourse-Aware Summarizer are taken directly from \cite{Cohan2018ADocuments}, while the results of Sent-CLF, Sent-PTR and TLM-I+E come from \cite{Subramanian2019OnModels}. Finally, results of PEGASUS and BigBird-PEGASUS are taken from \cite{zaheer2020big}.

\begin{table}[t!]
\caption{\label{results_arxiv} ROUGE F1 results on arXiv test set. Underlined are the top performing models in each category while bold is the overall top performing model.}
\begin{center}
\setlength\tabcolsep{0.1cm}
\begin{tabular}{rrccc}
\hline {\bf Model} & {\bf Type} & {\textbf{ROUGE-1}} & {\textbf{ROUGE-2}} & {\textbf{ROUGE-L}}\\\hline
SumBasic & Ext & 29.47 & 6.95 & 26.3\\
LexRank & Ext & 33.85 & 10.73 & 28.99\\
LSA & Ext & 29.91 & 7.42 & 25.67\\
Lead-10 & Ext & 35.52 & 10.33 & 31.44\\
Sent-CLF & Ext & 34.01 & 8.71 & 30.41\\
Sent-PTR & Ext & \underline{42.32} & \underline{15.63} & \underline{38.06}\\
Attention Seq2Seq & Abs & 29.3 & 6.00 & 25.56\\
PEGASUS & Abs & 44.21 & 16.95 & 38.83\\
BigBird-PEGASUS & Abs & \underline{\textbf{46.63}} & \underline{\textbf{19.02}} & \underline{\textbf{41.77}}\\
Pointer-Generator & Mix & 32.06 & 9.04 & 25.16\\
Discourse-Aware & Mix & 35.8 & 11.05 & 31.8\\
TLM-I+E & Mix & 42.43 & 15.24 & 24.08\\\hline
\multicolumn{5}{c}{\textbf{Our Models}}\\\hline
DANCER LSTM & Mix & 41.87 & 15.92 & 37.61\\
DANCER RUM & Mix & 42.7 & 16.54 & 38.44\\
DANCER PEGASUS & Abs & \underline{45.01} & \underline{17.60} & \underline{40.56}\\
\hline
\end{tabular}
\end{center}
\end{table}

\begin{table}[t!]
\caption{\label{results_pubmed} ROUGE F1 results on PubMed test set. Underlined are the top performing models in each category while bold is the overall top performing model.}
\begin{center}
\setlength\tabcolsep{0.1cm}
\begin{tabular}{rrccc}
\hline {\bf Model} & {\bf Type} & {\textbf{ROUGE-1}} & {\textbf{ROUGE-2}} & {\textbf{ROUGE-L}}\\\hline
SumBasic & Ext& 37.15 & 11.36 & 33.43\\
LexRank & Ext & 39.19 & 13.89 & 34.59\\
LSA & Ext & 33.89 & 9.93 & 29,70\\
Lead-10 & Ext & 37.45 & 14.19 & 34.07\\
Sent-CLF & Ext & \underline{45.01} & \underline{19.91} & \underline{41.16}\\
Sent-PTR & Ext & 43.3 & 17.92 & 39.47\\
Attention Seq2Seq & Abs & 31.55 & 8.52 & 27.38\\
PEGASUS & Abs & 45.97 & 20.15 & 41.34\\
BigBird-PEGASUS & Abs & \underline{46.32} & \underline{\textbf{20.65}} & \underline{42.33}\\
Pointer-Generator & Mix & 35.86 & 10.22 & 29.69\\
Discourse-Aware & Mix & 38.93 & 15.37 & 35.21\\
TLM-I+E & Mix & 41.43 & 15.89 & 24.32\\\hline
\multicolumn{5}{c}{\textbf{Our Models}}\\\hline
DANCER LSTM & Mix & 44.09 & 17.69 & 40.27\\
DANCER RUM & Mix & 43.98 & 17.65 & 40.25\\
DANCER PEGASUS & Abs & \underline{\textbf{46.34}} & \underline{19.97} & \underline{\textbf{42.42}}\\
\hline
\end{tabular}
\end{center}
\end{table}

\subsection{Transformer vs LSTM vs RUM}
Looking at the comparisons between the different DANCER variants we can see that the PEGASUS model is the clear winner. This is expected since it is a much more powerful and advanced model which makes use of extensive unsupervised pre-training. The price for the better performance is the increased requirements in terms of memory and processing power of PEGASUS compared to the simpler RNN models.

On the other hand both RNN models exhibit similar performance with the RUM model outperforming the LSTM model on the arXiv dataset, while performing slightly worse on PubMed. Given the observation that LSTM based models tend to copy more phrases from the source than RUM based models \cite{Dangovski2019RotationalApplications}, we hypothesize that the target abstracts in PubMed include a higher amount of text that is copied directly from the full text, compared to arXiv.

\begin{figure}
    \includegraphics[width=0.5\textwidth]{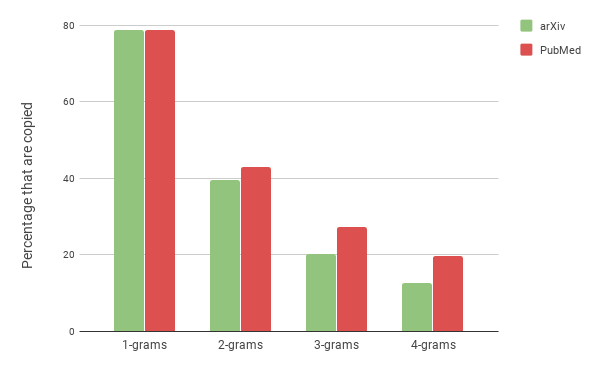}
    \caption{The percentage of N-grams that are copied directly from the source to the target summary for both datasets. The percentages are high for both datasets but for the PubMed dataset we observe a higher percentage of copied 2-grams, 3-grams, 4-grams. This implies that the abstracts of the articles are in fact very much extractive and as a result this dataset favors extractive approaches more.}
    \label{fig:copied_percent}
\end{figure}

In order to validate this hypothesis we computed the percentage of n-grams in the target summaries that are copied from the source. In Figure \ref{fig:copied_percent} we show these percentages for both datasets. It is clear that the target abstracts in the PubMed dataset have a greater percentage of copied 2-grams, 3-grams and 4-grams compared to the arXiv dataset. 

In addition, we found that when using a decoder with RUM units, the training is more stable than when using a decoder with LSTM units and converges steadily at a lower loss value. This is in line with the observation that RUM based models exhibit larger gradients and as a result have more robust training compared to LSTM based models \cite{Dangovski2019RotationalApplications}. On the other hand, we also found that models with a RUM based decoder need more steps to converge to the final loss, compared to models with an LSTM based decoder.

Overall, based on our experiments and analysis of the three different DANCER models, we conclude that the DANCER PEGASUS model is clearly superior to the RNN models. Nevertheless, when combined with DANCER both RNN models achieve a surprisingly good performance, although not as good as the PEGASUS model. This leads us to believe that there might still be some merit in using RNN sequence-to-sequence models especially in some low resource scenarios.

\subsection{DANCER vs baselines and the state of the art}

Based on the numbers of tables \ref{results_arxiv} and \ref{results_pubmed} we can see that DANCER works well with both the Pointer-Generator and the PEGASUS model. DANCER improves the performance of the Pointer-Generator model by almost 10 ROUGE-1 points which is a very significant improvement considering that the underlying model is the same. This model is also on par with models such as TLM-I+E as well as Sent-PTR despite using a significantly simpler architecture.

When combined with stronger models such as PEGASUS we can see that DANCER can still lead to improved results with minimal additional effort and resources. Moreover, we empirically show that DANCER can take advantage of strong pre-trained models, such as PEGASUS, and increase the effectiveness of task specific fine-tuning.

Our experiments show that DANCER PEGASUS is on par with the BigBird-PEGASUS model, which is the current state-of-the-art, without modifying the underlying model architecture of PEGASUS. With that in mind, it is possible that a combination of BigBird-PEGASUS with DANCER could further improve results although it was not in the scope of this work to explore that. Furthermore, more extensive optimization of the DANCER PEGASUS model, as well as additional training could lead to even better results but this was not the focus of this work. In general, the experimental results suggest that DANCER is a very easy to implement way to boost the performance of different summarization models with minimal additional effort and resources.




Going back to Figure \ref{fig:copied_percent}, we notice that both datasets have a high percentage of text copied directly from the source, which explains the high performance of all extractive approaches, even simple ones, like LexRank and Lead-10. Usually it is way easier for extractive models to achieve higher ROUGE scores due to the way that ROUGE metrics are calculated. Since the metric is purely based on the overlap of the the generated text with the target text and in many cases the target summary includes a parts that are copied from the source input, ROUGE scores clearly favor extractive summarization approaches. Nevertheless, we can see that advanced abstractive models such as PEGASUS, BigBird-PEGASUS and DANCER PEGASUS manage to outperform most extractive models by a significant margin. This is important since abstractive summarization is a more challenging task, but also more closely resembles the way humans do summarization.

In the Appendix of this paper we present sample summaries for a couple of papers generated by our models trained on the arXiv dataset. These samples demonstrate the quality of the summaries we can produce using our proposed methods as well as directly compare the outputs produced by the different summarization models.  

\section{Conclusion}
\label{sec:conclusion}

We presented DANCER, a novel summarization method for long documents. We focused on the summarization of academic articles, but the same method can easily be applied to different types of long documents, such as financial reports. We have demonstrated quantitatively through experiments on the arXiv and PubMed datasets that this method combined with a basic sequence-to-sequence RNN model can still achieve good performance. We also show that using a stronger model such as PEGASUS we can achieve results that are on par with the state-of-the-art on both datasets.

We have also evaluated the advantages of using a combination of LSTM and RUM units inside the sequence-to-sequence model in terms of ROUGE F1 as well as training stability and convergence. We have found that including RUM units in the decoder of the model can lead to a more stable training and better convergence as well as improved ROUGE scores, when the target sequence includes less text directly copied from the source sequence.

Overall, we have focused on the effectiveness of our proposed method regardless of the complexity of the core model. We emphasize that DANCER is a simple yet effective extension that can boost the performance of different summarization models with minimal additional effort and resources. In future work we would like to combine DANCER with more complex summarization models that could potentially further improve summarization quality as well as apply DANCER summarization on domains other than academic articles.

\bibliographystyle{IEEEtran}
%
\bibliography{dancer_summarization}


%

\onecolumn
\appendix[Examples of generated summaries]
In order to demonstrate the high quality of the abstracts produced by our models, we generated summaries from a couple of notable papers in our field. We are presenting examples generated from both the DANCER LSTM Pointer-Genarator and DANCER PEGASUS models as well as the ROUGE1-F1 scores for comparison. The models used to generate those summaries were trained on the arXiv dataset. We also provide the original abstract of each paper for reference and comparison purposes.

In the case of the first paper shown in Table \ref{table:bert} we see that although the PEGASUS model performs better in terms of ROUGE1 score, neither of the two models achieve very good results and both generated summaries are significantly different from the original abstract. On the other hand, for the second paper shown in Table \ref{table:nmt} we see that both models achieve high scores with the Pointer-Generator model outperforming the PEGASUS model.

\begin{table*}[h!]
\caption{\label{table:bert}}
\begin{center}
\setlength\tabcolsep{0.2cm}
\begin{tabularx}{\linewidth}{r|X}
\hline
\textbf{Title} & BERT: Pre-training of Deep Bidirectional Transformers for Language Understanding\\\hline
\textbf{Original abstract} & We introduce a new language representation model called BERT, which stands for Bidirectional Encoder Representations from Transformers. Unlike recent language representation models (Peters et al., 2018a; Radford et al., 2018), BERT is designed to pretrain deep bidirectional representations from unlabeled text by jointly conditioning on both left and right context in all layers. As a result, the pre-trained BERT model can be finetuned with just one additional output layer to create state-of-the-art models for a wide range of tasks, such as question answering and language inference, without substantial taskspecific architecture modifications. BERT is conceptually simple and empirically powerful. It obtains new state-of-the-art results on eleven natural language processing tasks, including pushing the GLUE score to 80.5\% (7.7\% point absolute improvement), MultiNLI accuracy to 86.7\% (4.6\% absolute improvement), SQuAD v1.1 question answering Test F1 to 93.2 (1.5 point absolute improvement) and SQuAD v2.0 Test F1 to 83.1 (5.1 point absolute improvement)\\\hline
\textbf{Pointer-Generator summary} & Language model pre-training has been shown to be effective for improving many natural language processing tasks (Dai and Le, 2015; Radford et al., 2018). The major limitation is that standard language models are unidirectional, and this limits the choice of architectures that can be used during pre-training. We argue that current techniques restrict the power of the pre-trained representations, especially for the fine-tuning approaches. The BERT model is trained on unlabeled data over different pre-training tasks. For finetuning, the BERT model is first initialized with the pre-trained parameters, and all of the parameters are fine-tuned using labeled data from the downstream tasks. These results enable the same pre-trained model to successfully tackle a broad set of NLP tasks. In particular, these findings enable even low-resource tasks to benefit from deep unidirectional architectures. Our major contribution is further generalizing these findings to deep bidirectional architectures.\\\hline
\textbf{ROUGE1-F1} & 35.0\\\hline
\textbf{PEGASUS summary} & Language model pre-training has been shown to be effective for improving many natural language processing tasks such as sentence-level paraphrasing and entity recognition tasks. However, current approaches to pre-trained language models are restricted to unidirectional language models. In this paper, we propose a new approach to pre-trained language models based on bidirectional encoder transformers (BERT). BERT is inspired by the pre-training objective of cloze task (Taylor et al., 1953), where the goal is to predict some masked language representations from the input. We introduce BERT and its detailed implementation in this paper. The BERT model is first initialized with the pre-trained parameters, and all of the parameters are fine-tuned using labeled data from the downstream tasks. Rich unsupervised pre-training is an integral part of many language understanding systems. In particular, these results enable even low-resource tasks to benefit from deep unidirectional architectures. Our major contribution is further generalizing these findings to deep bidirectional architectures, allowing the same pre-trained model to successfully tackle a broad set of NLP tasks.\\\hline
\textbf{ROUGE1-F1} & 36.52\\\hline
\end{tabularx}
\end{center}
\end{table*}

\begin{table*}[t!]
\caption{\label{table:nmt}}
\begin{center}
\setlength\tabcolsep{0.2cm}
\begin{tabularx}{\linewidth}{r|X}
\hline
\textbf{Title} & Neural Machine Translation by Jointly Learning to Align and Translate\\\hline
\textbf{Original abstract} & Neural machine translation is a recently proposed approach to machine translation. Unlike the traditional statistical machine translation, the neural machine translation aims at building a single neural network that can be jointly tuned to maximize the translation performance. The models proposed recently for neural machine translation often belong to a family of encoder–decoders and encode a source sentence into a fixed-length vector from which a decoder generates a translation. In this paper, we conjecture that the use of a fixed-length vector is a bottleneck in improving the performance of this basic encoder–decoder architecture, and propose to extend this by allowing a model to automatically (soft-)search for parts of a source sentence that are relevant to predicting a target word, without having to form these parts as a hard segment explicitly. With this new approach, we achieve a translation performance comparable to the existing state-of-the-art phrase-based system on the task of English-to-French translation. Furthermore, qualitative analysis reveals that the (soft-)alignments found by the model agree well with our intuition.\\\hline
\textbf{Pointer-Generator summary} & Neural machine translation is a newly emerging approach to machine translation, recently proposed by Kalchbrenner and Blunsom (2013), Sutskever et al. (2014) and Cho et al., (2003) which consists of the encoder and the decoder for a language pair that are jointly trained to maximize the probability of a correct translation given a source sentence. We show that the proposed approach of jointly learning to align and translate achieves significantly improved translation performance. The performance of the RNNsearch is as high as that of the conventional phrase-based translation system (Moses), when only the sentences consisting of known words are considered. This is a significant achievement, considering that Moses uses a separate monolingual corpus (418m words) in addition to the parallel corpora we used to train the RNNsearch and RNNencdec. One of the motivations behind the proposed approach was the basic encoder-decoder approach to underperform with long sentences. We show that the proposed approach provides an intuitive way to inspect the (soft-) alignment between the words in a generated translation and those in a source sentence. This is done by visualizing the annotation weights. In this paper, we propose a novel approach to neural machine translation, called an encoder-decoder approach, encodes a whole input sentence into a fixed-length vector from which a translation will be decoded. We conjectured that the proposed RNNsearch outperforms the conventional encoder-decoder model (RNNencdec) significantly, regardless of the sentence length and that it is much more robust to the length of a source sentence.\\\hline
\textbf{ROUGE1-F1} & 54.17\\\hline
\textbf{PEGASUS summary} & Neural machine translation is a newly emerging approach to machine translation, recently proposed by Kalchbrenner and Blunsom, Sutskever et al. and Cho et al. The proposed RNNsearch outperforms the conventional RNNencdec when only the sentences consisting of known words are considered. More importantly, the performance of the RNNsearch is as high as that of the conventional phrase-based translation system (Moses) when only the sentences consisting of known words are considered. The proposed approach provides an intuitive way to inspect the (soft-)alignment between the words in a generated translation. This is done by visualizing the weights associated with the annotation of the source sentence and those associated with the annotation of the target word. The conventional approach to neural machine translation, called an encoder-decoder approach, encodes a whole input sentence into a fixed-length context vector from which a translation will be decoded. We conjectured that the use of a fixed-length context vector is problematic for translating long sentences, based on a recent empirical study reported by Cho et al.\\\hline
\textbf{ROUGE1-F1} & 52.12\\\hline
\end{tabularx}
\end{center}
\end{table*}

\end{document}